\documentclass{article} 
\usepackage{iclr2027_conference,times}

\usepackage{amsmath,amsfonts,bm}

\def\eqref#1{equation~\ref{#1}}

\def\1{\bm{1}}

\DeclareMathAlphabet{\mathsfit}{\encodingdefault}{\sfdefault}{m}{sl}
\SetMathAlphabet{\mathsfit}{bold}{\encodingdefault}{\sfdefault}{bx}{n}

\usepackage{etoolbox}

\usepackage{hyperref}
\usepackage{url}
\usepackage{bbm}
\usepackage{pifont}

\usepackage{xcolor}
\usepackage{tcolorbox}

\definecolor{linkbg}{HTML}{F3F7FB}
\definecolor{linkborder}{HTML}{D9E5F0}
\definecolor{linkblue}{HTML}{356A91}

\definecolor{tablegroup}{RGB}{244,247,251}
\definecolor{tablescaleFour}{RGB}{232,240,249}
\definecolor{tablescaleNine}{RGB}{232,244,239}
\definecolor{tablebest}{RGB}{218,239,226}
\definecolor{vadfigurenavy}{HTML}{14213D}
\definecolor{vadfiguregreen}{HTML}{08783E}
\definecolor{vadfigurepalegreen}{HTML}{F5FAF7}

\usepackage{microtype}
\microtypesetup{expansion=false}
\usepackage{amsmath}
\usepackage{amssymb}
\usepackage{mathtools}
\usepackage[T1]{fontenc}
\usepackage[utf8]{inputenc}
\usepackage{array}
\usepackage{colortbl}
\usepackage{tabularx}
\usepackage{enumitem}
\usepackage{float}
\usepackage[percent]{overpic}
\usepackage{makecell}
\usepackage{balance}
\usepackage{fontawesome5}
\usepackage{booktabs}
\usepackage{algorithm}
\usepackage{algpseudocode}
\usepackage{dashrule}
\usepackage{dashrule}
\usepackage{graphicx}
\usepackage{arydshln}
\usepackage{float}

\newcommand{\bestscore}[1]{{\bfseries\boldmath #1}}
\newcommand{\secondscore}[1]{\underline{#1}}
\newcommand{\panelref}[2]{\hyperref[#1]{\ref*{#1}#2}}

\title{CV-OPD: Contrastive Vision On-policy Distillation}

\author{
\mbox{%
Haowei Wang$^{1,2}$\thanks{These authors contributed equally to this work.}%
\hspace{0.5em}
Hao Lin$^{1}$\footnotemark[1]\hspace{0.5em}
Qi Zhu$^{2}$\hspace{0.5em}
Jiayi Ji$^{1}$\hspace{0.5em}
Deyi Ji$^{2}$\hspace{0.5em}
Liujuan Cao$^{1}$\hspace{0.5em}
Rongrong Ji$^{1}$
}
}

\affiliation{
$^{1}$Key Laboratory of Multimedia Trusted Perception and Efficient Computing,
Ministry of Education of China, Xiamen University\\
$^{2}$Moxin Technology
}

\email{
\texttt{\{ishaowei.wang\}@gmail.com}
}

\usepackage{amsthm}
\newtheorem{observation}{Observation}

\iclrfinalcopy 

\begin{document}
\thispagestyle{empty}
\pagestyle{plain} 

\maketitle
\begin{abstract}
On-policy distillation (OPD) reveals versatility and convenience during VLM post-training, providing dense supervision for student from teacher.
However, we observed an anomaly that the distilled VLM could still answer the questions accurately even when the related visual regions were masked or replaced with noise.
Building on this observation, our quantitative and qualitative experiments demonstrate that OPD, which uses the answer as a hint, has failed to teach students to genuinely seek out and rely on visual evidence. 
Our theoretical analysis suggests that this failure arises from a \textit{shortcut enabled by the lack of visual constraints} during distillation.
To address the issue, we introduce \textbf{C}ontrastive \textbf{V}ision \textbf{O}n-\textbf{p}olicy \textbf{D}istillation (CV-OPD), which compares teacher distributions induced by positive and negative crops while conditioning on the same student prefix and augments optimization with a margin-based contrastive constraint, which is modulated with discrepancy gating across token and sample level.
Specifically, CV-OPD first introduces a negative teacher unrelated to the answer and designs a Contrastive Divergence Loss (CDL) that constrains the feature distribution of student by contrasting the positive and negative teachers, correcting how the student exploits visual regions. 
The CV-OPD introduces a Discrepancy Aware Gate (DAG) that uses token-level and sample-level discrepancies to select visually relevant tokens for optimization, directing the student’s attention toward visual content.
Extensive experiments across a broad range of fine-grained visual benchmarks demonstrate that CV-OPD substantially alleviates shortcut and consistently improves the visual reasoning capabilities of existing VLMs.

\begin{center}
\begin{tcolorbox}[
    colback=linkbg,
    colframe=linkborder,
    boxrule=0.4pt,
    arc=2mm,
    left=3mm,
    right=3mm,
    top=1.5mm,
    bottom=1.5mm,
    width=0.92\linewidth,
    before skip=2mm,
    after skip=3mm,
    halign=left,
]
\small
\textbf{Code:}\,
\href{https://github.com/HaoLin3419/Contrastive-Vision-OPD}{
    \textcolor{linkblue}{https://github.com/HaoLin3419/Contrastive-Vision-OPD}
}
\\
\textbf{Project:}\,
\href{https://haolin3419.github.io/Contrastive-Vision-OPD}{
    \textcolor{linkblue}{https://haolin3419.github.io/Contrastive-Vision-OPD}
}
\end{tcolorbox}
\end{center}

\end{abstract}

\begin{figure}[h]
\begin{center}
\includegraphics[width=\textwidth]{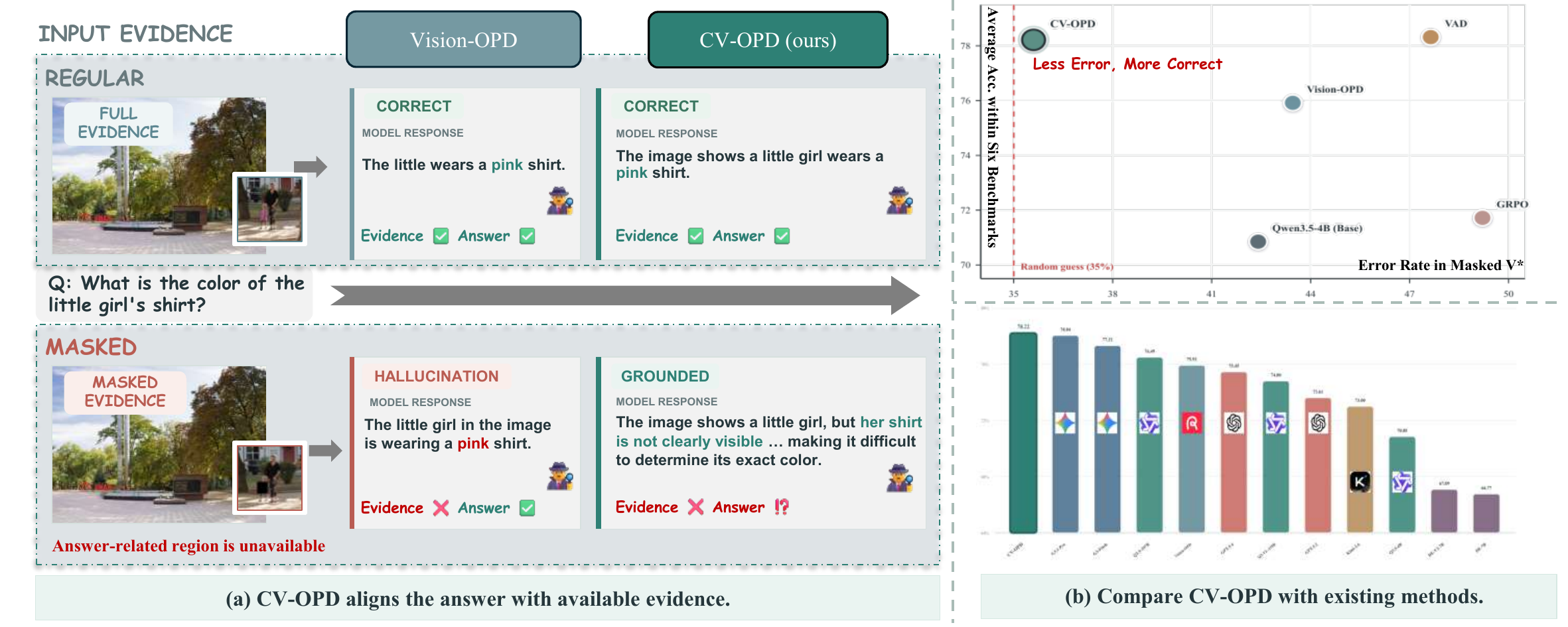}
\end{center}
\caption{(a) The failure mode of Vision-OPD. The distilled VLM can generate correct answer even given wrong visual inputs. (b) Our CV-OPD shows less error with higher accuracy, which is even competitive while compared to closed-source VLMs.}
\label{fig1}
\end{figure}

\section{Introduction}

In recent LLM post-training, on-policy distillation (OPD)~\citep{agarwal2024policy} has gained increasing attention as a means of addressing the lack of token-level supervision in reinforcement learning~\citep{shao2024deepseekmath}. 
By training the student to match the teacher’s output distribution over its own on-policy generations, OPD effectively transfers the teacher’s capabilities while alleviating the distribution mismatch between training and inference. 
A common practice is to equip the teacher with privileged information~\citep{liu2026visualadvantage,yuan2026vision}, enabling it to identify and correct errors in the student’s responses. 
These properties make OPD particularly well suited to handling the challenging samples encountered in Vision Language Model (VLM) post-training~\citep{yuan2026vision,sun2026v}.

Naturally, recent works, \emph{e.g.}, Vision-OPD~\citep{yuan2026vision}, VA-OPD~\citep{liu2026visualadvantage} and so on, have extended OPD to VLMs by providing the teacher with cropped or enlarged target regions to correct erroneous reasoning trajectories of the student, yielding substantial performance gains. 
However, as shown on Fig.~\ref{fig1}, we uncover a significant issue of the distilled VLMs.
In a simple diagnostic experiment, we mask the image regions relevant to the answer and evaluate the distilled model. 

Surprisingly, Vision-OPD still generates correct answers without necessary visual evidence, achieving 43.46\% and 48.69\% accuracy on $\text{V}^*$~\citep{wu2024vstar} under black masking and Gaussian noise respectively, substantially exceeding the random rate for multiple-choice questions.
This result suggests that directly exposing answer-relevant information to the teacher induces a \textbf{visual shortcut}, through which the VLM can bypass the underlying visual evidence to produce answers.
In Sec.~\ref{sec:observation}, we examine this phenomenon in detail and provide a formal derivation of its possible causes.

To address this limitation, we propose \textbf{C}ontrastive \textbf{V}ision \textbf{O}n-\textbf{p}olicy \textbf{D}istillation (CV-OPD), which constructs negative visual samples to explicitly discourage reliance on misleading regions.
By contrasting positive and negative visual evidence, CV-OPD encourages the model to learn not only \textbf{what answer to produce}, but also \textbf{where to look and what to avoid}, thereby refining the solution space of response generation.

Specifically, CV-OPD introduces a \textbf{negative teacher} receives a selected image region that is unrelated with answer, and generates distractive content conditioned on the student’s sampled trajectory. 
The difference between two teachers naturally captures model biases and style-related shortcuts that are irrelevant to the visual evidence. 
We then introduce a Contrastive Divergence Loss (CDL), which explicitly steers the student’s token distribution away from the negative teacher with hinge margin.
CDL encourages the model to not merely imitate the trajectory to answer  the question, but to identify and reject visually irrelevant evidence. 
Finally, we propose a Discrepancy Aware Gate (DAG) to select token sequences that are highly relevant to the reasoning process. The constraint is applied only to these selected tokens, directing the optimization towards useful visual contents.

We evaluate CV-OPD on a broad range of fine-grained visual benchmarks following the previous work~\citep{yuan2026vision}, where it consistently delivers substantial improvements over vanilla OPD. More importantly, experiments in masked $\text{V}^*$ show that CV-OPD is significantly more grounded in visual evidence.

In summary, our contributions can be concluded into three-fold:

\begin{itemize}
    \item We uncover a previously overlooked shortcut phenomenon in OPD for VLMs and demonstrate that it primarily arises from directly exposing answer-relevant visual information to the teacher.

    \item We propose \textbf{C}ontrastive \textbf{V}ision \textbf{O}n-\textbf{p}olicy \textbf{D}istillation (CV-OPD), which leverages negative teachers constructed from randomly sampled visual regions. The proposed CDL and DAG encourage the model to learn not only the correct answer, but also an evidence-grounded reasoning trajectory for identifying its visual source.

    \item CV-OPD consistently improves performance across a broad range of visual benchmarks comapred with the base model. Furthermore, experiments in masked $\text{V}^*$ demonstrate that CV-OPD substantially strengthens the reliance on visual evidence for VLMs.
\end{itemize}
\section{Observation and Analysis}
\label{sec:observation}
\subsection{Preliminary}

On-policy distillation~\citep{agarwal2024policy} provides dense supervision by leveraging a stronger teacher to reshape the token-level output distributions along trajectories sampled by the student. 
In VLM post-training, this paradigm typically equips the teacher with privileged visual hints to identify and correct errors in the student’s generated responses.
For each sample, giving the original visual content $V$ and question $Q$, student first sample the $t$-th token $\pi_t(k)$ as:
\begin{equation}
    \pi_t(k)=
\pi_\theta(A_t=k\mid V,C_t),
\label{eq:student}
\end{equation}
where $C_t$ is the context including $V$ and sampled prefix $A_{<t}$ and $A_t$ means the $t$-th token in output answer. $k$ is the token index of the dictionary.
Compared with student, the difference of privileged teacher is the extra visual hint of inputs. The distribution of teacher can be formulated as:
\begin{equation}
    \tau_t^+(k)=
\tau_\phi(A_t=k\mid C_t,H^+),
\label{eq:teacher}
\end{equation}
while the $H^+$ is the hint information, which is cropped or zoom-in region related with answers.
Based on that, the standard optimization target of OPD in all data $\mathcal{D}$ can be formulated as:
\begin{equation}
\mathcal L_{\mathrm{OPD}}(\theta)
=
\mathbb E_{\mathcal D}
\mathbb E_{A_{1:T}\sim\pi_{\bar\theta}}
\left[
\sum_{t=1}^{T}w_t
D_{\mathrm{JS}}(\pi_t,\tau_t^+)
\right].
\label{eq:opd}
\end{equation}
The rollout distribution and teacher distribution are treated as stop-gradient during the optimization and only the parameters $\theta$ of student is optimized.

\subsection{Observations}

\begin{figure}[t]
    \centering
    \includegraphics[width=\linewidth]{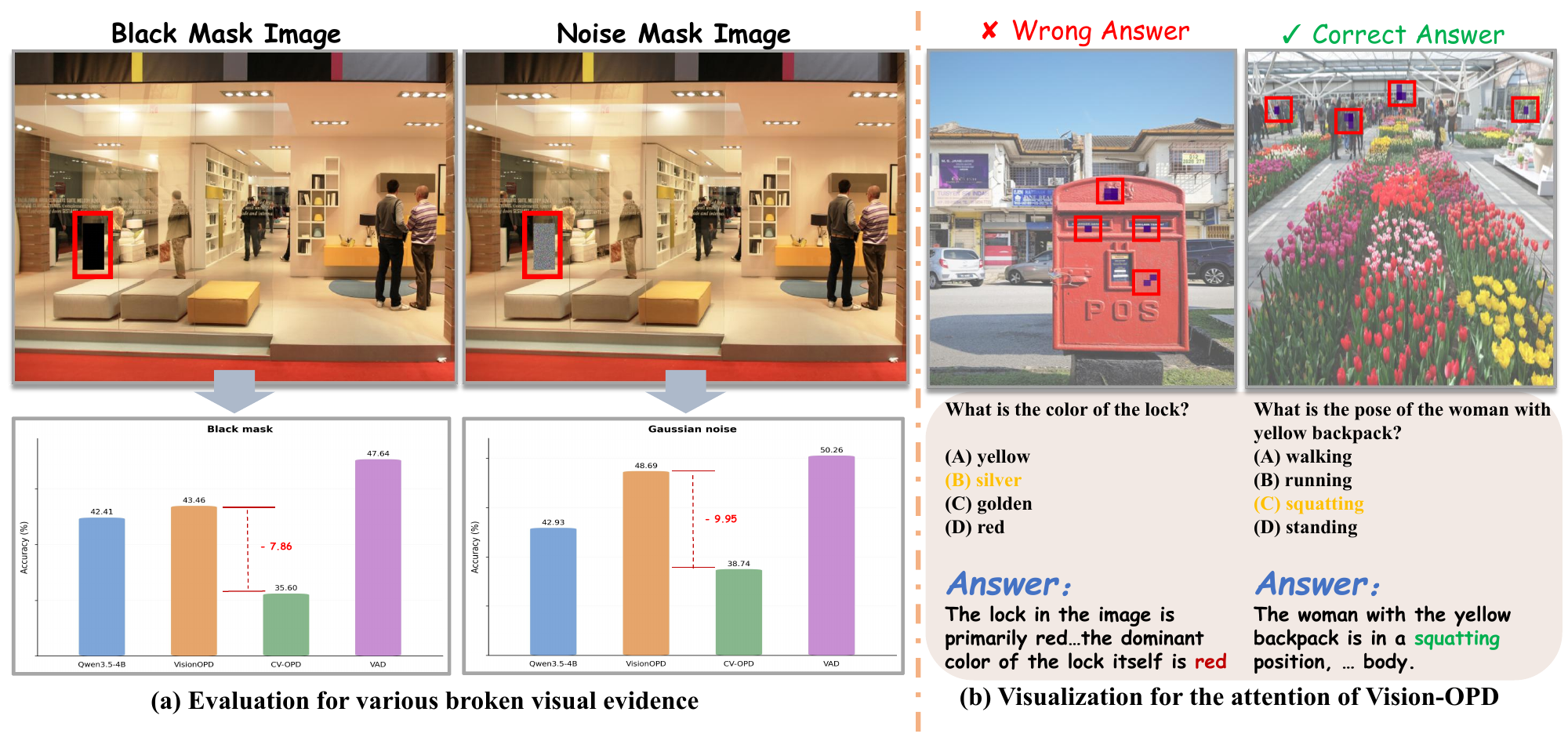}

    \caption{
        Diagnostic evaluation of visual shortcut.
        (a) We evaluate accuracy under both black masking and Gaussian noise,
        where answer-relevant visual regions are replaced.
        (b) We further visualize the attention (\textcolor{red}{red rectangle}) of Vision-OPD to find out whether the correct visual information is utilized.
    }

    \label{fig:findings}
\end{figure}

Although OPD appears promising, we identified the following observations in the context of VLMs.
\begin{observation}
    The distilled VLM can handle questions without visual evidence.
\end{observation}
We replace the answer-relevant regions in the $\text{V}^*$~\citep{wu2024vstar} with black pixels or Gaussian noise and ensure no extra interference from ambiguity, then evaluate Vision-OPD~\citep{yuan2026vision} on the processed images. As shown in Fig.~\ref{fig:findings} (a), Vision-OPD achieves
$43.46\%$ and $48.69\%$ accuracy under black masking and
Gaussian noise, respectively, exceeding the base model's
$42.41\%$ and $42.93\%$.
Despite the removal of answer-relevant visual information,
the distilled model becomes more accurate at predicting
the answers. These results suggest that visual self-distillation may introduce shortcut learning~\citep{geirhos2020shortcut}, allowing the model to produce correct answers without properly grounding in visual evidence.

\begin{observation}
    The attention of distilled VLM is unrelated with vision content.
\end{observation}
We visualize the attention of self-distilled Vision-OPD both on the correct and wrong situation.
As shown in Fig.~\ref{fig:findings} (b),
the model fails to consistently focus on the target regions
in both correctly and incorrectly answered examples.
Even a correct prediction does not necessarily mean that
the model has identified the relevant visual evidence.
This reveals a gap between answer accuracy and visual
grounding in self-distilled VLMs. More visualizations and analysis can be found in~\ref{app:attention} and ~\ref{app:black-mask-attention}.

\subsection{Analysis}
These observations naturally lead us to a more fundamental question, \textbf{what drives the emergence of such shortcut during distillation?} We therefore formalize the distillation process and revisit its underlying optimization objective to identify the source of this behavior.

First, define $H^+$ as the hint for positive teacher. It's obviously the mutual information between $H^+$ and answer is higher than original image. According to the optimization target Eq.~\ref{eq:opd}, due to the symmetry of JSD~\citep{lin1991divergence}, what the student learns is the conditional Fréchet Mean of the privileged teacher. Therefore, student retains the same policy of privileged teacher token posterior:
\begin{equation}
    \arg\max_k \pi_{t,k}^*\approx\arg\max_k\tau_{t,k}^{+}.
\end{equation}

Based on this, the gradient of JSD with respect to the student logits $z_t$ can be expressed as:

\begin{equation}
\nabla_{z_t}D_{\mathrm{JS}}(\pi_t,\tau_t^+)
=
\frac{1}{2}
\left(
\operatorname{Diag}(\pi_t)
-
\pi_t\pi_t^\top
\right)
\log
\frac{2\pi_t}{\pi_t+\tau_t^+}.
\label{eq:gradient}
\end{equation}
It's obviously that JSD will increase the logit of hint related target token. 
At the same time, the student distribution is decided with visual content $V$, question $Q$, and prefix $A_{<t}$ as shown on Eq.~\ref{eq:student}. 
Such gradients impose no constraint on what the student should optimize.
Compared with the target from Eq.~\ref{eq:teacher}, because the mapping from $V$ to $H^+$ is neither differentiable nor observable, optimization instead resorts to the full image and the fixed textual context. 
Consequently, the model learns to generate outputs by fitting the available conditions rather than actively identifying positive region $R^+$.

\section{Contrastive Vision On-policy Distillation}

\begin{figure}[!htbp]
\begin{center}
\includegraphics[width=\textwidth]{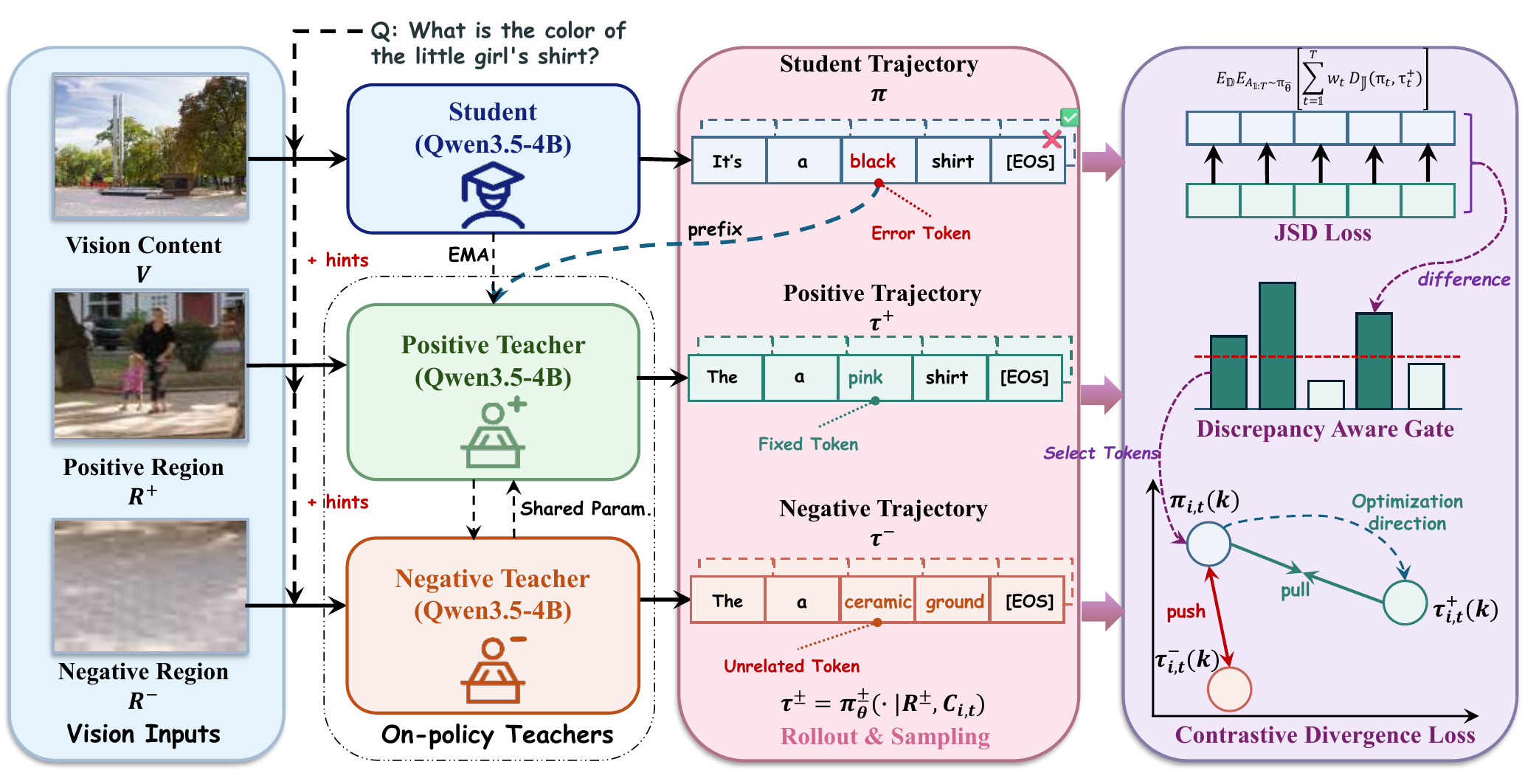}
\end{center}
\caption{The pipeline of Contrastive Vision On-policy Distillation. CV-OPD introduces a negative teacher that provides unrelated visual information to fix the reliance of visual contents. Samples with at least one correct rollout are retained for training.}
\label{fig2}
\end{figure}

To mitigate the shortcut during distillation, we propose Contrastive Vision On-policy Distillation (CV-OPD). As the crown jewels, CV-OPD introduces a \textbf{negative teacher} that receives randomly sampled visual regions irrelevant to the answer, and employs a Contrastive Divergence Loss (CDL) and Discrepancy Aware Gate (DAG) to encourage the model to distinguish visual evidence from irrelevant cues. 

\subsection{Overview}

As shown in Fig.~\ref{fig2}, given the vision inputs and text prompt, for each example, a frozen copy of the current student, $\pi_{\theta}$, generates $M$ on-policy trajectories from the same image:
\begin{equation}
A_{1:T_i}^{(i)}\sim
\prod_{t=1}^{T_i}
\pi_{\theta}
\!\left(A_t^{(i)}\mid V,Q,A_{<t}^{(i)}\right),
\qquad
C_{i,t}=(Q,A_{<t}^{(i)}),
\quad i=1,\ldots,M.
\end{equation}
Constructed on the student trajectory, the positive region $R^+$ is drawn from an answer related proposal. Besides, we construct an extra pipeline for the negative region $R^-$, which is sampled from the same image, with its height and width matched to the positive region and its overlap constrained by a threshold $\delta$:
\begin{equation}
    \begin{aligned}
        R^+&\sim\rho^+(\cdot\mid V,Q), \\
        R^-&\sim\rho^-(\cdot\mid V,Q,R^+),\quad\text{where }     \mathrm{IoU}\!\left(R^+,R^-\right)\leq\delta.
    \end{aligned}
\end{equation}
The $\rho$ is the proposal mechanism to select the image region related to the conditions.  
The teacher branches share parameters and operate on the same prefix from student, differing only in the privileged crop.
For the $t$-th token, the resulting teacher token distributions are:
\begin{equation}
    \tau^+_{i, t}=\pi^+_\theta(\cdot|R^+, C_{i, t}), \tau^-_{i, t}=\pi^-_\theta(\cdot|R^-, C_{i, t})
\end{equation}
For the positive teacher distribution $\tau^+$ and student distribution $\pi$, we rewrite the Jensen-Shannon divergence as the optimized target as shown on Eq.~\ref{eq:opd} with the average across all the rollouts from student. As for the negative teacher, we would take its advantages in the following sections.

\subsection{Contrastive Divergence Loss}

With the negative teacher in place, we introduce a contrastive divergence loss that explicitly encourages the model to capture the distributional differences induced by distinct visual prefixes, rather than indiscriminately aligning with the teacher’s output.

Specifically, at each position $(i,t)$, we compute the full-vocabulary Jensen--Shannon divergence between the student and each teacher distribution. The resulting positive and negative divergences are compared through a margin-based hinge loss:

\begin{equation}
    \widetilde d_{i,t}^{\pm}
=D_{\mathrm{JS}}\!\left(\pi_{i,t},\tau_{i,t}^{\pm}\right),
\qquad
\mathcal L_{\mathrm{CDL}}(\theta)
=\mathbb E_{\mathcal D}
\mathbb E_{A_{1:T}\sim \pi_{\theta}}
\left[\sum_{i=1}^{M}w_t(
\mu+\widetilde d_{i,t}^{+}-\widetilde d_{i,t}^{-})
\right]_+.
\label{eq:CDL}
\end{equation}
Here, $[x]_+=\max(x,0)$ and $0<\mu<\log 2$. Minimizing $\mathcal L_{\mathrm{CDL}}(\theta)$ draws the student toward the positive teacher while pushing it away from the negative teacher. Once $\widetilde d_{i,t}^{-}-\widetilde d_{i,t}^{+}\geq\mu$, the loss drops to zero, so the negative branch is not pushed away indefinitely.

In CDL, set $F(\cdot)$ is the Fisher Information Matrix in Eq.~\ref{eq:gradient} and $u(\cdot, \cdot)$ is the log part. The gradient with respect to the student logits $z_t$ can be expressed as:
\begin{equation}
    \Delta z_t
\propto
F(\pi_t)
\left[ u(\pi_t,\tau_t^-)
-
u(\pi_t,\tau_t^+)
\right].
\label{eq:CDL_gradient}
\end{equation}
The exact form of the gradient is:
\begin{equation}
    -\nabla_{z_t}
\mathcal L_{CDL}(\theta)
=
\frac12F(\pi_t)
\log
\frac{\pi_t+\tau_t^+}
{\pi_t+\tau_t^-},
\label{eq:11}
\end{equation}
which is proportional to $\tau^+_t-\tau_t^-$. This finding indicates that CDL can itself amplify the sampling discrepancy between the positive and negative inputs, thereby helping the model learn to localize positive regions. The detailed proof of gradient can be viewed in~\ref{app:prove_CDL}

\subsection{Discrepancy Aware Gate}

Go further, to alleviate gradient dilution arising from sampling noise and stylistic variations~\citep{liang2026visual}, we seek to concentrate the visual content related optimization signals of CV-OPD on tokens that are most relevant to the underlying visual evidence.

Therefore, we introduce Discrepancy Aware Gate (DAG), using the full-vocabulary divergence between student and positive teacher as the token-level evidence score to retain the useful tokens for the gradient backpropagation later within the batch:
\begin{equation}
\bar d_{\mathcal B}
=
\frac{1}{|\mathcal T_{\mathcal B}|}
\sum_{(i,t)\in\mathcal T_{\mathcal B}}
d_{i,t},
\qquad
g_{i,t}
=
\mathbf 1
\left[
d_{i,t}>\bar d_{\mathcal B}
\right].
\label{eq:discrepancy_gate}
\end{equation}
Put simply, the contrastive term kicks in only at positions whose positive discrepancy exceeds the batch average. The gate $g_{i,t}$ selects sequence positions.

Beyond token-level filtering, we further consider the reliability of teacher supervision. Since the teacher samples conditioned on the student’s prefix, the student’s correctness directly influences the likelihood of obtaining a valid teacher correction, making low-quality student trajectories a potential source of noisy supervision.
Let $r_i\in\{0,1\}$ indicate whether the final answer from rollout $i$ is correct. We define a sample-level reliability gate as

\begin{equation}   
b=\mathbf1\!\left[\sum_{i=1}^{M}r_i>0\right].
\end{equation}

An incorrect rollout can leave the teacher conditioning on an off-track prefix. If this happens for all $M$ rollouts, we disable all the supervision term for that example.

\subsection{Training Objective}

Let $w_{i,t}\geq0$ with $\sum_t w_{i,t}=1$, and let $\lambda\geq0$ weight the CDL term. Combining standard distillation and gated CDL gives

\begin{equation}
\mathcal{L}(\theta)
=
\mathbb{E}_{\mathcal{D}}
\left[
\frac{b}{M}
\sum_{i=1}^{M}
\sum_{t=1}^{T_i}
\left(
w_{i,t}^{\mathrm{JSD}} d_{i,t}
+
\lambda g_{i,t}
w_{i,t}^{\mathrm{CDL}}
\left[
\mu
+
\widetilde{d}_{i,t}^{+}
-
\widetilde{d}_{i,t}^{-}
\right]_{+}
\right)
\right].
\label{eq:training_objective}
\end{equation}

The expectation covers training examples, positive and negative region sampling, and student rollouts. Since $b=\mathbf1[\sum_i r_i>0]$, distillation is enabled only when at least one rollout is correct. In practice, we use $w_{i,t}=1/T_i$. 
\section{Experiments}
\subsection{Setup}

We train CV-OPD on Qwen3.5-4B~\citep{qwenteam2026qwen354b} using 6.2K visual reasoning samples from Vision-OPD~\citep{yuan2026vision}. The primary distillation objective is Jensen--Shannon Divergence (JSD) with $\beta=0.5$, computed over the Top-100 vocabulary entries and an aggregated tail bucket. An Exponential Moving Average (EMA) teacher tracks the student parameters at an update rate of 0.05. Each sample generates eight on-policy rollouts of up to 1024 tokens, sharing one negative crop for contrastive supervision. Samples with no correct rollouts are excluded from distillation.

Evaluation covers six fine-grained visual benchmarks: $\text{V}^*$~\citep{wu2024vstar}, ZoomBench~\citep{wei2026zooming}, HR-Bench~\citep{wang2025divide} 4K and 8K, and MME-RealWorld~\citep{zhang2024mme} EN and CN. All Qwen3.5-based models operate in non-thinking mode, and GPT-OSS-120B~\citep{openai2025gptoss} serves as the answer judge.

\subsection{Main Results}

\begin{table*}[!t]
  \centering
  \small
  \setlength{\tabcolsep}{1.5pt}
  \renewcommand{\arraystretch}{1.08}
  \renewcommand{\tabularxcolumn}[1]{m{#1}}

  \caption{Overall comparison on six fine-grained visual benchmarks. We report the models trained on the same data within the Qwen3.5-Based Models. Values are accuracies (\%), and $\operatorname{Avg}_6$ is the unweighted mean. Within the Qwen3.5-based block, best results are bold and second-best results are underlined; the cyan-green row denotes CV-OPD.}

  \label{tab:main-results}

  \begin{tabularx}{\textwidth}{
    @{}
    >{\raggedright\arraybackslash}m{4.9cm}
    >{\centering\arraybackslash}m{0.65cm}
    *{7}{>{\centering\arraybackslash}X}
    @{}
  }

  \toprule

    {\footnotesize Model} &
    {\footnotesize Param.} &
    {\footnotesize V$^\star$} &
    {\footnotesize Zoom} &
    \shortstack{\scriptsize HR-Bench\\[-1pt]\footnotesize 4K} &
    \shortstack{\scriptsize HR-Bench\\[-1pt]\footnotesize 8K} &
    \shortstack{\scriptsize MME-RW\\[-1pt]\footnotesize EN} &
    \shortstack{\scriptsize MME-RW\\[-1pt]\footnotesize CN} &
    {\footnotesize $\operatorname{Avg}_6$} \\

    \midrule

    \rowcolor{tablegroup}
    \multicolumn{9}{c}{\itshape Tool-Using and Region-Aware Models} \\

    DeepEyes~{\scriptsize\citep{zheng2026deepeyes}}
& 7B & $82.63$ & $46.15$ & $75.50$ & $70.00$
& $63.95$ & $62.41$ & $66.77$ \\

SenseNova-MARS~{\scriptsize\citep{chng2026sensenovamars}}
& 8B & $92.15$ & $47.81$ & $83.13$ & $78.38$
& $67.90$ & $68.90$ & $73.05$ \\

Vision-RL$^2$~{\scriptsize\citep{shi2026visionrl2}}
& 7B & $91.62$ & $59.88$ & $78.12$ & $75.00$
& $62.41$ & $58.73$ & $70.96$ \\

\midrule

\rowcolor{tablegroup}
\multicolumn{9}{c}{\itshape Closed-Source Models} \\

Gemini 3 Flash~{\scriptsize\citep{googledeepmind2025gemini3flash}}
& -- & $85.39$ & $61.49$ & $87.25$ & $85.00$
& $73.15$ & $71.66$ & $77.32$ \\

Gemini 3.1 Pro~{\scriptsize\citep{googledeepmind2026gemini31pro}}
& -- & $88.48$ & $62.01$ & $88.12$ & $84.50$
& $73.32$ & $71.81$ & $78.04$ \\

GPT-5.2~{\scriptsize\citep{openai2025gpt52}}
& -- & $84.59$ & $54.71$ & $84.64$ & $78.48$
& $71.82$ & $67.43$ & $73.61$ \\

GPT-5.4~{\scriptsize\citep{openai2026gpt54}}
& -- & $85.83$ & $57.94$ & $86.15$ & $78.10$
& $74.16$ & $70.50$ & $75.45$ \\

\midrule

\rowcolor{tablegroup}
\multicolumn{9}{c}{\itshape Open-Source Models} \\

Qwen3.5-397B-A17B~{\scriptsize\citep{qwenteam2026qwen35}}
& 397B & $84.23$ & $58.30$ & $91.25$ & $84.37$
& $74.17$ & $66.64$ & $76.49$ \\

Kimi-K2.5~{\scriptsize\citep{kimiteam2026kimik25}}
& 1T & $85.86$ & $56.33$ & $81.87$ & $75.38$
& $71.51$ & $68.40$ & $73.23$ \\

Qwen3-VL-Instruct~{\scriptsize\citep{bai2025qwen3vltechnicalreport}}
& 8B & $84.82$ & $42.96$ & $79.63$ & $75.25$
& $63.19$ & $64.61$ & $68.41$ \\

\midrule

\rowcolor{tablegroup}
\multicolumn{9}{c}{\itshape Qwen3.5-Based Models} \\

Qwen3.5~{\scriptsize\citep{qwenteam2026qwen354b}}
& 4B
& $82.20$
& $48.40$
& \bestscore{$86.38$}
& \secondscore{$82.00$}
& $63.94$
& $63.80$
& $71.12$ \\

GRPO~{\scriptsize\citep{shao2024deepseekmath}}
& 4B
& $81.68$
& $56.57$
& $76.75$
& $75.62$
& $71.10$
& $68.58$
& $71.72$ \\

VAD~{\scriptsize\citep{zhang2026vad}}
& 4B
& \bestscore{$91.62$}
& $60.12$
& \secondscore{$85.13$}
& \bestscore{$83.38$}
& \secondscore{$76.98$}
& \bestscore{$71.44$}
& \secondscore{$78.11$} \\

VA-OPD~{\scriptsize\citep{liu2026visualadvantage}}
& 4B
& $86.91$
& \secondscore{$60.36$}
& $81.62$
& $78.12$
& $73.29$
& $67.42$
& $74.62$ \\

V-Zero~{\scriptsize\citep{sun2026v}}
& 4B
& $88.48$
& $56.45$
& $84.88$
& $78.75$
& $73.49$
& $70.45$
& $75.42$ \\

Vision-OPD~{\scriptsize\citep{yuan2026vision}}
& 4B
& $89.53$
& $59.41$
& $81.75$
& $80.12$
& $74.51$
& $70.17$
& $75.92$ \\

Decomposed OPD~{\scriptsize\citep{yoon2026decomposed}}
& 4B
& \secondscore{$89.80$}
& $58.70$
& $82.60$
& $79.50$
& $73.80$
& $67.82$
& $75.37$ \\

\rowcolor{tablebest}
\textbf{CV-OPD (Ours)}
& 4B
& \bestscore{$91.62$}
& \bestscore{$62.01$}
& $84.38$
& \bestscore{$83.38$}
& \bestscore{$76.99$}
& \secondscore{$70.93$}
& \bestscore{$78.22$} \\
    \bottomrule

\end{tabularx}
\end{table*}
As shown on Tab.~\ref{tab:main-results}, most notably, CV-OPD consistently improves performance across domains when built upon the Qwen-3.5 backbone, demonstrating strong competitiveness against both open- and closed-source foundation models. Compared with Vision-OPD, CV-OPD improves the average performance across metrics by 2.3\%. On the Zoom dataset, it consistently outperforms existing methods by approximately 2\%. These results demonstrate that CV-OPD can attend to relevant visual regions more accurately and remains competitive with widely used proprietary models. Notably, on HR-Bench 4K, all distillation methods underperform the base model, which we attribute to advantages in the original pretraining for 4K inputs. \textbf{Regarding training cost}, the primary additional computational overhead comes from the negative teacher. However, enabled by KV caching and the design for sharing teacher, CV-OPD increases training time by only approximately 10\% relative to the original Vision-OPD.

\subsection{Ablation Study}

We have conducted controlled ablations to understand the contribution of each component in CV-OPD. Specifically, we investigate three questions: (1) how to construct the pipeline, (2) how to construct the negative teacher, and (3) how to control the gate.

\subsubsection{Q1: How to Construct Pipeline}
\begin{table*}[!t]
  \centering
  \small
  \setlength{\tabcolsep}{1.5pt}
  \renewcommand{\arraystretch}{1.08}
  \renewcommand{\tabularxcolumn}[1]{m{#1}}

  \caption{Ablation of CDL and DAG in the CV-OPD pipeline.}
  \label{tab:pipeline-ablation}

  \begin{tabularx}{\textwidth}{
    @{}
    >{\centering\arraybackslash}m{0.9cm}
    >{\centering\arraybackslash}m{0.9cm}
    *{7}{>{\centering\arraybackslash}X}
    @{}
  }

    \toprule

    \multicolumn{2}{c}{\footnotesize Method}
    & \multicolumn{7}{c}{\footnotesize Evaluation} \\

    \midrule

    {\footnotesize CDL} &
    {\footnotesize DAG} &
    {\footnotesize V$^\star$} &
    {\footnotesize Zoom} &
    \shortstack{\scriptsize HR-Bench\\[-1pt]\footnotesize 4K} &
    \shortstack{\scriptsize HR-Bench\\[-1pt]\footnotesize 8K} &
    \shortstack{\scriptsize MME-RW\\[-1pt]\footnotesize EN} &
    \shortstack{\scriptsize MME-RW\\[-1pt]\footnotesize CN} &
    {\footnotesize $\operatorname{Avg}_6$} \\

    \midrule

    \ding{55} & \ding{55}
    & 89.53 & 59.41 & 81.75 & 80.12
    & 74.51 & 70.17 & 75.92 \\

    \ding{55} & \ding{52}
    & 90.05 & 60.71 & 82.38 & 82.63
    & 74.67 & 70.41 & 76.81 \\

    \ding{52} & \ding{55}
    & 90.58 & 60.71 & 83.00 & 80.88
    & 75.40 & 70.71 & 76.88 \\

    \rowcolor{tablebest}
    \ding{52} & \ding{52}
    & 91.62 & 62.01 & 84.38 & 83.38
    & 76.99 & 70.93 & \textbf{78.22} \\

    \bottomrule

  \end{tabularx}

\end{table*}

We examine the contributions of the Contrastive Divergence Loss (CDL) and the Discrepancy Aware Gate (DAG).
As shown in Table~\ref{tab:pipeline-ablation}, removing CDL reduces $\operatorname{Avg}_6$ from $78.22\%$ to $76.81\%$.
The contrast between positive and negative teachers helps the student distinguish answer-relevant visual evidence from irrelevant cues.
Without DAG, the average drops to $76.88\%$, suggesting that informative token selection provides more effective supervision than treating all positions equally. The variant without either component achieves only $75.92\%$, while their combination reaches $78.22\%$.
These results highlight the importance of precise visual evidence supervision and its alignment with informative tokens in the distillation pipeline.

\subsubsection{Q2: How to Select the Negative Teacher}
\begin{table*}[!t]
  \centering
  \small
  \setlength{\tabcolsep}{1.5pt}
  \renewcommand{\arraystretch}{1.08}
  \renewcommand{\tabularxcolumn}[1]{m{#1}}

  \caption{Ablation of negative region selection strategies}
  \label{tab:teacher-ablation}

  \begin{tabularx}{\textwidth}{
    @{}
    >{\raggedright\arraybackslash}m{3.5cm}
    >{\centering\arraybackslash}m{0.65cm}
    *{7}{>{\centering\arraybackslash}X}
    @{}
  }

    \toprule

    {\footnotesize Model} &
    {\footnotesize Param.} &
    {\footnotesize V$^\star$} &
    {\footnotesize Zoom} &
    \shortstack{\scriptsize HR-Bench\\[-1pt]\footnotesize 4K} &
    \shortstack{\scriptsize HR-Bench\\[-1pt]\footnotesize 8K} &
    \shortstack{\scriptsize MME-RW\\[-1pt]\footnotesize EN} &
    \shortstack{\scriptsize MME-RW\\[-1pt]\footnotesize CN} &
    {\footnotesize $\operatorname{Avg}_6$} \\

    \midrule

    Base & 4B
    & 82.20 & 48.40 & 86.38 & 82.00
    & 63.94 & 63.80 & 71.12 \\

    All Black & 4B
    & 91.62 & 59.76 & 81.88 & 81.62
    & 74.94 & 70.73 & 76.76 \\

    \midrule

    Random Crop & 4B
    & 89.53 & 62.01 & 84.38 & 81.50
    & 75.06 & 70.73 & 77.20 \\

    Neighbor Crop & 4B
    & 90.05 & 59.76 & 81.62 & 81.00
    & 75.40 & 70.54 & 76.40 \\

    \rowcolor{tablebest}
    \textbf{CV-OPD} & 4B
    & 91.62 & 62.01 & 84.38 & 83.38
    & 76.99 & 70.93 & \textbf{78.22} \\

    \bottomrule

  \end{tabularx}

\end{table*}

Besides CV-OPD, we explore three alternative strategies for constructing the negative teacher: an all-black image, a crop neighboring the target region, and a randomly sampled crop. In Random Crop, each rollout independently samples a negative region, while CV-OPD shares the same negative crop across all eight rollouts of each sample.

As shown in Table~\ref{tab:teacher-ablation}, Neighbor Crop achieves an $\operatorname{Avg}_6$ of $76.40\%$, below the variant without CDL ($76.81\%$). Regions near the target may still contain answer-relevant visual information, which can introduce conflicting supervision when the student is pushed away from the negative teacher. All Black achieves $76.76\%$, while Random Crop improves the average to $77.20\%$. CV-OPD further raises the performance to $78.22\%$. The performance gap between Random Crop and CV-OPD may stem from their different negative sampling strategies. By sharing the same negative crop across eight rollouts, CV-OPD reduces variation in contrastive gradients and provides more consistent supervision for optimization.

\subsubsection{Q3: How to Control the Gate}
\begin{table*}[!t]
  \centering
  \small
  \setlength{\tabcolsep}{1.5pt}
  \renewcommand{\arraystretch}{1.08}
  \renewcommand{\tabularxcolumn}[1]{m{#1}}

  \caption{Ablation of DAG threshold settings.}
  \label{tab:gate-ablation}

  \begin{tabularx}{\textwidth}{
    @{}
    >{\centering\arraybackslash}m{1.0cm}
    >{\centering\arraybackslash}m{1.5cm}
    *{7}{>{\centering\arraybackslash}X}
    @{}
  }

    \toprule

    {\footnotesize DAG} &
    {\footnotesize Threshold} &
    {\footnotesize V$^\star$} &
    {\footnotesize Zoom} &
    \shortstack{\scriptsize HR-Bench\\[-1pt]\footnotesize 4K} &
    \shortstack{\scriptsize HR-Bench\\[-1pt]\footnotesize 8K} &
    \shortstack{\scriptsize MME-RW\\[-1pt]\footnotesize EN} &
    \shortstack{\scriptsize MME-RW\\[-1pt]\footnotesize CN} &
    {\footnotesize $\operatorname{Avg}_6$} \\

    \midrule

    \ding{55} & --
    & 90.58 & 60.71 & 83.00 & 80.88
    & 75.40 & 70.71 & 76.88 \\

    \ding{52} & rollout
    & 90.05 & 61.54 & 83.25 & 82.13
    & 75.05 & 70.47 & 77.08 \\

    \rowcolor{tablebest}
    \ding{52} & batch
    & 91.62 & 62.01 & 84.38 & 83.38
    & 76.99 & 70.93 & \textbf{78.22} \\

    \bottomrule

  \end{tabularx}

\end{table*}

We compare two thresholding strategies for DAG, using either the mean discrepancy of each rollout or the mean discrepancy across the batch.
As shown in Table~\ref{tab:gate-ablation}, the rollout-level threshold achieves an $\operatorname{Avg}_6$ of $77.08\%$, only slightly above the variant without DAG ($76.88\%$).
For short responses, the average discrepancy can be sensitive to a few token positions, making it difficult to reliably identify tokens that require fine-grained contrastive supervision.
Using the batch-level mean instead raises $\operatorname{Avg}_6$ to $78.22\%$.
The shared threshold provides a more consistent reference across rollouts, allowing DAG to better capture informative tokens and direct contrastive supervision toward them. So a robust threshold is essential for controlling the gate, as it remains resilient to extreme cases and ensures stable, accurate token selection.
\section{Related Work}
\label{sec:related_work}

\subsection{On-Policy Distillation in Large Language Models}

On-policy distillation (OPD) trains a student model to match the teacher's distribution on its own generated trajectories. GKD~\citep{agarwal2024policy} demonstrates that this approach can reduce the distribution mismatch between training sequences and student outputs at inference time, thereby mitigating exposure bias. Prior work has explored student-generated training contexts, alternative divergence objectives, and efficient sample reuse to improve autoregressive distillation~\citep{lin2020autoregressive,gu2023minillm,ko2024distillm}. On-policy self-distillation (OPSD) further removes the need for an external teacher by deriving supervision from the student itself. For example, SDFT~\citep{shenfeld2026self} conditions a self-teacher on expert demonstrations to provide stronger on-policy supervision. However, direct access to reference solutions can introduce answer leakage~\citep{wei2026amr}, while the provided context can encourage shortcut-like rationalization and harm subsequent reasoning distillation~\citep{lee2026answer}. These findings highlight the importance of how privileged information is used in distillation.

\subsection{On-Policy Distillation in Vision-Language Models}

Recent studies have extended OPD to vision-language models by providing teachers with privileged visual information~\citep{zhang2026vad,liu2026visualadvantage,yoon2026decomposed}. VA-OPD~\citep{liu2026visualadvantage} emphasizes visually informative tokens through visual advantage, enabling more targeted distillation of fine-grained visual knowledge. From a different perspective, Decomposed OPD~\citep{yoon2026decomposed} disentangles language-prior and visual-grounding gradients to strengthen the model's reliance on visual evidence. Other approaches explore contrastive visual evidence to improve the quality of distillation supervision. V-Zero~\citep{sun2026v} uses paired positive and negative visual views to assess the evidence supporting student-generated trajectories and gate token-level distillation. VAD~\citep{zhang2026vad} further investigates the visual attribution of teacher corrections by comparing evidence-present and evidence-removed views, reconstructing supervision targets from visually attributable distributional changes. These studies highlight the importance of distinguishing visual evidence from language priors during distillation. In particular, Vision-OPD~\citep{yuan2026vision} uses an evidence-centered crop-conditioned teacher to supervise an original-image-conditioned student through token-level distillation. This regional-to-global formulation transfers the benefits of focused visual inputs to original-image inference. However,this approach also raises the question of whether students have learned to identify visual evidence independently. Building on this setting, we investigate how crop-guided supervision affects students' reliance on visual evidence and introduce contrastive teacher supervision to mitigate shortcut learning.


\section{Conclusion}
In this work, we introduced \textbf{C}ontrastive \textbf{V}ision \textbf{O}n-\textbf{p}olicy \textbf{D}istillation (CV-OPD) to mitigate optimization shortcuts induced by the direct exposure of answer-relevant information during distillation. CV-OPD incorporates a randomly sampled visual negative teacher and a margin-based \textbf{Contrastive Divergence Loss} (CDL) to encourage the model to distinguish relevant visual evidence from misleading cues and to learn evidence-grounded reasoning trajectories. In addition, CV-OPD further employs a Discrepancy Aware Gate (DAG) that combines token-level distributional discrepancies with sample-level reliability filtering to improve the precision of distillation supervision. These designs enable CV-OPD to achieve substantial improvements in both cross-domain generalization and hallucination mitigation, demonstrating the effectiveness of contrastive supervision for promoting visually grounded reasoning in VLMs.

\bibliography{iclr2027_conference}
\bibliographystyle{iclr2027_conference}

\appendix
\section{Appendix}

\subsection{Detailed Derivation of CDL}
\label{app:prove_CDL}
Start from the Eq.~\ref{eq:CDL_gradient}, the gradient of $k$-th token is:
\begin{equation}
    \begin{aligned}
\frac{\partial\mathcal L_{CDL}}
{\partial \pi_t(k)}
&=
\frac12
\log\frac{2\pi_t(k)}{\pi_t(k)+\tau_t^+(k)}
-
\frac12
\log\frac{2\pi_t(k)}{\pi_t(k)+\tau_t^-(k)}
\\
&=
\frac12
\log
\frac{\pi_t(k)+\tau_t^-(k)}
{\pi_t(k)+\tau_t^+(k)},
\end{aligned}
\label{eq:15}
\end{equation}
which is the former format of the Eq.~\ref{eq:11}. It reveals that the optimization direction of the token $t$ relies on the difference between positive distribution and negative distribution. In the local geometric case, the update direction degenerates to $\tau_t^+ - \tau_t^-$.

We define the $\pi_t$, $\tau_t^+$, and $\tau_t^-$ belong to the neighbor domain of the latent distribution $r_t$ like:
\begin{equation}
    \pi_t=r_t+\delta_t^\pi,
\qquad
\tau_t^+=r_t+\delta_t^+,
\qquad
\tau_t^-=r_t+\delta_t^-.
\end{equation}
When the jitter norm of the perturbation is small, the local second-order expansion of JSD is:
\begin{equation}
    D_{\mathrm{JS}}(\pi_t,\tau_t)
=
\frac18
(\pi_t-\tau_t)^\top
\operatorname{Diag}(r_t)^{-1}
(\pi_t-\tau_t)
+
\mathcal O(\|\pi_t-\tau_t\|_2^3).
\end{equation}
Therefore, the gradient in Eq.~\ref{eq:11} can be rewrite as:
\begin{equation}
    \nabla_{z_t}
\mathcal L_{CDL}
=
\frac14(\tau_t^- - \tau_t^+)
+
\mathcal O(\|\delta_t\|_2^2),
\end{equation}
which is proportional to $\tau_t^+ - \tau_t^-$.

Original JSD uses the teacher's absolute distribution as supervision. As for CDL, under a first-order approximation, CDL uses the difference in teacher distributions resulting from matched crop pairs as supervision.

\subsection{Training Hyperparameters}
\label{app:hyperparameters}

Our training settings are summarized in Table~\ref{tab:training-hyperparameters}.
CV-OPD is trained on eight A100 GPUs using Qwen3.5-4B, a batch size of 96, and eight on-policy rollouts per sample.
The maximum prompt and response lengths are 8192 and 1024 tokens, respectively.
Training runs for one epoch under AdamW optimization at a learning rate of $2\times10^{-6}$.
The primary distillation objective is generalized JSD ($\beta=0.5$), computed over Top-100+tail probability distributions.
An EMA teacher tracks the student parameters at an update rate of 0.05.
For contrastive distillation, both the margin and loss weight are set to 0.1.
Negative crops are shared across rollouts of the same sample, subject to a maximum IoU of 0.1.

\subsection{Training Algorithm}
\label{app:algorithm}

The complete training procedure of CV-OPD is summarized in
Algorithm~\ref{alg:cv-opd}. At each training step, the student generates
multiple on-policy rollouts from the full image, while the EMA teacher
provides positive and negative crop-conditioned distributions along
the same prefix. A shared negative crop is used for all rollouts of
each sample, and samples with no correct rollouts are excluded from
distillation. The remaining trajectories are optimized using the main
JSD distillation objective and the gated Contrastive Divergence Loss
(CDL), followed by an EMA update of the teacher parameters.

\subsection{Attention Visualization}
\label{app:attention}

\begin{figure}[H]
    \centering

    \begingroup
    \setlength{\fboxsep}{4pt}
    \setlength{\fboxrule}{0.5pt}

    \fbox{%
        \includegraphics[
            width=0.94\linewidth,
            height=0.30\textheight,
            keepaspectratio
        ]{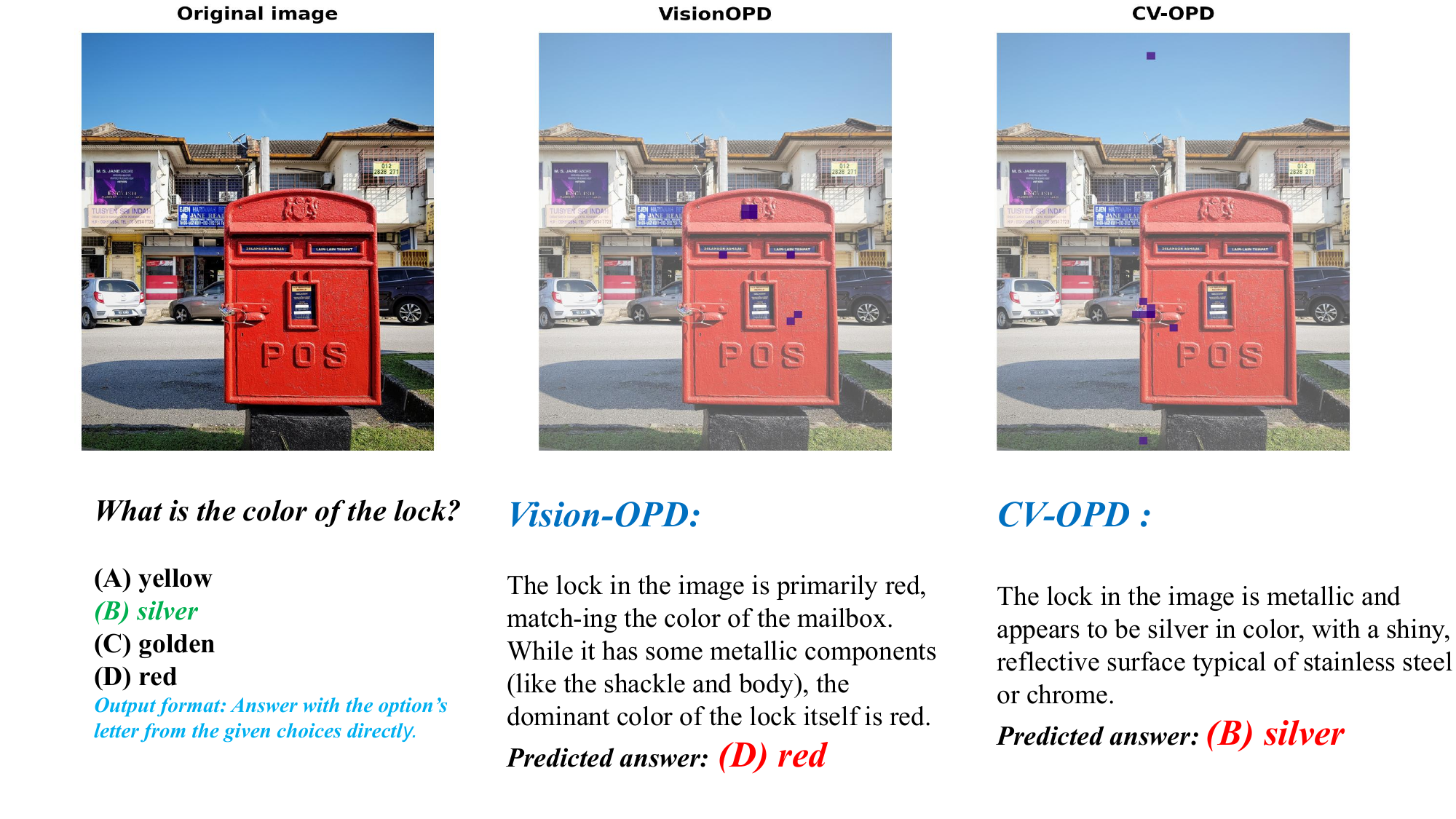}%
    }

    \endgroup

    \caption{
        Qualitative comparison of visual attention and
    model responses between Vision-OPD and CV-OPD.
    Vision-OPD relies on the dominant color of the
    mailbox and predicts red, whereas CV-OPD focuses
    on the lock itself and identifies its metallic
    silver color.
    }

    \label{fig:attention-case1}

\end{figure}

To examine whether CV-OPD can more accurately localize
answer-relevant visual regions, we compare its attention
distributions with those of Vision-OPD on fine-grained
visual reasoning examples.
As shown in Fig.~\ref{fig:attention-case1},
CV-OPD exhibits more concentrated attention on
answer-relevant regions, while Vision-OPD shows relatively
dispersed attention across the image.
Based on the results, CV-OPD directs more attention to the
visual details needed to answer the question.
This tighter focus improves fine-grained visual understanding
and reduces reliance on irrelevant cues and shortcut-based
predictions.


\begin{table}[!htbp]
  \centering
  \footnotesize
  \setlength{\tabcolsep}{4pt}
  \renewcommand{\arraystretch}{1.18}

  \caption{Training hyperparameters for CV-OPD.}
  \label{tab:training-hyperparameters}

  \begin{tabularx}{\linewidth}{
    @{}
    >{\raggedright\arraybackslash}p{0.29\linewidth}
    >{\raggedright\arraybackslash}p{0.25\linewidth}
    >{\raggedright\arraybackslash}X
    @{}
  }

    \toprule
    \textbf{Setting} & \textbf{Value} & \textbf{Role} \\
    \midrule

    GPU allocation
    & 8 A100 GPUs
    & Training worker allocation \\

    Backbone
    & Qwen3.5-4B
    & Student and EMA teacher model \\

    Training samples
    & 6,241
    & Number of visual reasoning examples \\

    Training batch size
    & 96
    & Number of samples per training batch \\

    Actor update mini-batch size
    & 96
    & Mini-batch size for actor updates \\

    Rollouts per prompt
    & 8
    & On-policy responses generated per sample \\

    Maximum prompt length
    & 8192 tokens
    & Prompt truncation limit \\

    Maximum response length
    & 1024 tokens
    & On-policy generation limit \\

    Maximum model length
    & 9216 tokens
    & Maximum sequence length \\

    Training epochs
    & 1
    & Number of passes over the training set \\

    Learning rate
    & $2\times10^{-6}$
    & Actor optimization learning rate \\

    Optimizer
    & AdamW
    & Student parameter optimization \\

    AdamW betas
    & $(0.9, 0.999)$
    & Optimizer momentum coefficients \\

    Weight decay
    & $0.01$
    & Optimizer regularization \\

    Gradient clipping
    & $1.0$
    & Maximum gradient norm \\

    Learning rate scheduler
    & Constant
    & Learning rate schedule \\

    Warmup steps
    & 10
    & Learning rate warmup \\

    Distillation objective
    & Generalized JSD
    & Main token-level distillation loss \\

    JSD coefficient
    & $0.5$
    & Main distillation interpolation coefficient \\

    Top-$K$ size
    & 100
    & Student-selected vocabulary support \\

    Tail probability bucket
    & Enabled
    & Aggregation of remaining vocabulary mass \\

    Distillation temperature
    & $1.0$
    & Temperature for the main distillation loss \\

    Teacher parameter update
    & EMA of student weights
    & Teacher model update strategy \\

    Teacher EMA update rate
    & $0.05$
    & Student-weight coefficient in EMA updates \\

    Positive teacher image field
    & \texttt{bbox\_images}
    & Source of answer-relevant crops \\

    Negative crop sharing
    & Sample-wise
    & One negative crop shared across eight rollouts \\

    Maximum negative crop IoU
    & $0.1$
    & Overlap constraint with the positive region \\

    Contrastive loss
    & Full-vocabulary JSD
    & Positive-negative teacher comparison \\

    Contrastive margin
    & $0.1$
    & Margin in the contrastive objective \\

    Contrastive loss weight
    & $0.1$
    & Weight of the auxiliary contrastive loss \\

    Token gate threshold
    & Batch-mean JSD
    & Selection of tokens for contrastive supervision \\

    Sample-level filter
    & All-wrong filtering
    & Exclusion of samples with eight incorrect rollouts \\

    Rollout correction mode
    & Token-level
    & Importance sampling correction \\

    Rollout correction threshold
    & $2.0$
    & Importance ratio clipping threshold \\

    GRPO loss weight
    & $0.0$
    & No additional GRPO policy loss \\

    Checkpoint frequency
    & 10 steps
    & Interval between saved actor checkpoints \\

    Logging mode
    & Console, TensorBoard, W\&B
    & Training metric recording \\

    \bottomrule

  \end{tabularx}
\end{table}

\clearpage

\begin{algorithm}[t]
\caption{CV-OPD Training}
\label{alg:cv-opd}

\begin{algorithmic}[1]

\Require
Training set $\mathcal{D}$, student policy $\pi_{\theta}$,
rollout count $M$, Top-$K$ size $K$,
JSD coefficient $\beta$, negative crop IoU threshold $\delta$,
contrastive margin $\mu$, contrastive weight $\lambda$,
EMA update rate $\tau$

\Ensure
Updated student parameters $\theta$

\State Initialize teacher parameters $\phi \gets \theta$

\For{each training step}

    \State Sample a mini-batch $\mathcal{B} \subset \mathcal{D}$
    \State Initialize retained trajectories $\mathcal{T} \gets \emptyset$

    \For{each sample $x=(I,q,R^{+},a)\in\mathcal{B}$}

        \State Generate $M$ on-policy rollouts
        $Y^{(1)},\ldots,Y^{(M)} \sim \pi_{\theta}(\cdot\mid I,q)$

        \State Compute correctness rewards
        $r_i \gets \operatorname{Reward}(Y^{(i)},a)$

        \If{$\sum_{i=1}^{M} r_i = 0$}
            \State \textbf{continue}
        \EndIf

        \State Sample one negative crop $R^{-}$ from $I$
        with $\operatorname{IoU}(R^{+},R^{-}) \leq \delta$

        \For{$i=1,\ldots,M$}

            \State Compute student distributions $\pi_{i,t}$
            using the full image $I$

            \State Compute positive teacher distributions
            $\tau^{+}_{i,t}$ using $R^{+}$

            \State Compute negative teacher distributions
            $\tau^{-}_{i,t}$ using the shared $R^{-}$

            \State Compute full-vocabulary discrepancies
            $d_{i,t} \gets D_{\mathrm{JS}}
            (\pi_{i,t},\tau^{+}_{i,t})$

            \State Add valid response-token positions to $\mathcal{T}$

        \EndFor

    \EndFor

    \If{$\mathcal{T}=\emptyset$}
        \State \textbf{continue}
    \EndIf

    \State Compute batch-mean discrepancy
    $\bar d \gets
    \frac{1}{|\mathcal{T}|}
    \sum_{(i,t)\in\mathcal{T}} d_{i,t}$

    \State Initialize losses
    $\mathcal{L}_{\mathrm{OPD}}\gets 0$,
    $\mathcal{L}_{\mathrm{CDL}}\gets 0$

    \State Initialize normalization factors
    $N\gets 0$, $G\gets 0$

    \For{each valid token position $(i,t)\in\mathcal{T}$}

        \State Construct student-aligned Top-$K$+tail
        distributions $\tilde{\pi}_{i,t}$ and
        $\tilde{\tau}^{+}_{i,t}$

        \State $\mathcal{L}_{\mathrm{OPD}}
        \gets \mathcal{L}_{\mathrm{OPD}}
        + w_{i,t}D_{\mathrm{JS},\beta}
        (\tilde{\pi}_{i,t},\tilde{\tau}^{+}_{i,t})$

        \State $N\gets N+w_{i,t}$

        \If{$d_{i,t}>\bar d$}

            \State $d^{+}_{i,t}\gets
            D_{\mathrm{JS}}(\pi_{i,t},\tau^{+}_{i,t})$

            \State $d^{-}_{i,t}\gets
            D_{\mathrm{JS}}(\pi_{i,t},\tau^{-}_{i,t})$

            \State $\mathcal{L}_{\mathrm{CDL}}
            \gets \mathcal{L}_{\mathrm{CDL}}
            +[\mu+d^{+}_{i,t}-d^{-}_{i,t}]_{+}$

            \State $G\gets G+1$

        \EndIf

    \EndFor

    \State $\displaystyle
    \mathcal{L}\gets
    \frac{\mathcal{L}_{\mathrm{OPD}}}{N+\epsilon}
    +\lambda
    \frac{\mathcal{L}_{\mathrm{CDL}}}{G+\epsilon}$

    \State Update student parameters $\theta$
    by minimizing $\mathcal{L}$

    \State Update teacher parameters
    $\phi\gets(1-\tau)\phi+\tau\theta$

\EndFor

\end{algorithmic}
\end{algorithm}

\subsection{Attention Visualization in Black Mask}
\label{app:black-mask-attention}

\begin{figure}[H]
    \centering

    \begingroup
    \setlength{\fboxsep}{4pt}
    \setlength{\fboxrule}{0.5pt}

    \fbox{%
        \includegraphics[
            width=0.94\linewidth,
            height=0.30\textheight,
            keepaspectratio
        ]{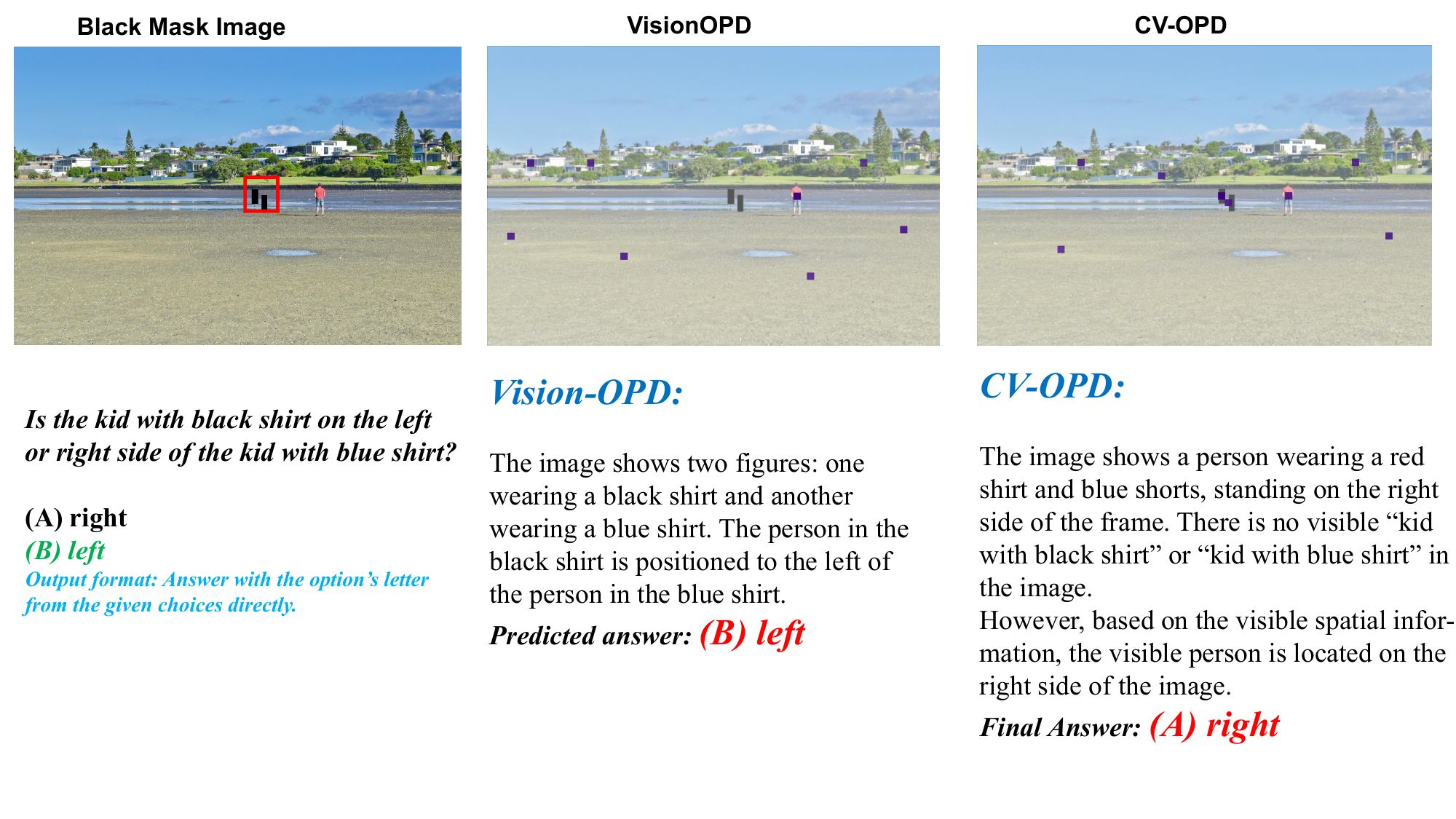}%
    }

    \endgroup

    \caption{
       Attention visualization and model responses under the black-mask setting.
    After the answer-relevant region is replaced by black pixels, Vision-OPD
    still produces the expected answer despite failing to localize the relevant
    visual evidence, indicating shortcut-based prediction and hallucination.
    In contrast, CV-OPD explicitly describes the observable scene and its
    ambiguity, showing stronger reliance on available visual evidence.
    }

    \label{fig:black-mask-case}

\end{figure}

We analyzed the shortcut behavior of Vision-OPD under the
black-mask setting, where the target region is replaced by
black pixels.
As shown in Fig.~\ref{fig:black-mask-case},
Vision-OPD still gives the correct answer even though the
relevant visual evidence is occluded.
This suggests that the model can rely on hallucinated or
shortcut-based reasoning rather than truly grounding its
prediction in the target region.
The attention visualization further supports this observation:
Vision-OPD does not accurately localize the correct target
position, yet it still predicts the expected answer.
In contrast, CV-OPD gives a more objective response by
explicitly describing the visible scene and the ambiguity
in the image, which indicates stronger reliance on
observable visual evidence instead of unsupported inference.

\subsection{Responses Under Insufficient Visual Evidence}
\label{app:insufficient-evidence}

We additionally present qualitative cases where answer-relevant
visual evidence is occluded or corrupted.
Despite the loss of visual information, CV-OPD explicitly
acknowledges the missing evidence and describes the observable
scene without inventing unsupported visual details.



\clearpage


\begin{figure}[H]
    \centering

    \begingroup
    \setlength{\fboxsep}{4pt}
    \setlength{\fboxrule}{0.5pt}

    \fbox{%
        \includegraphics[
            width=0.94\linewidth,
            height=0.30\textheight,
            keepaspectratio
        ]{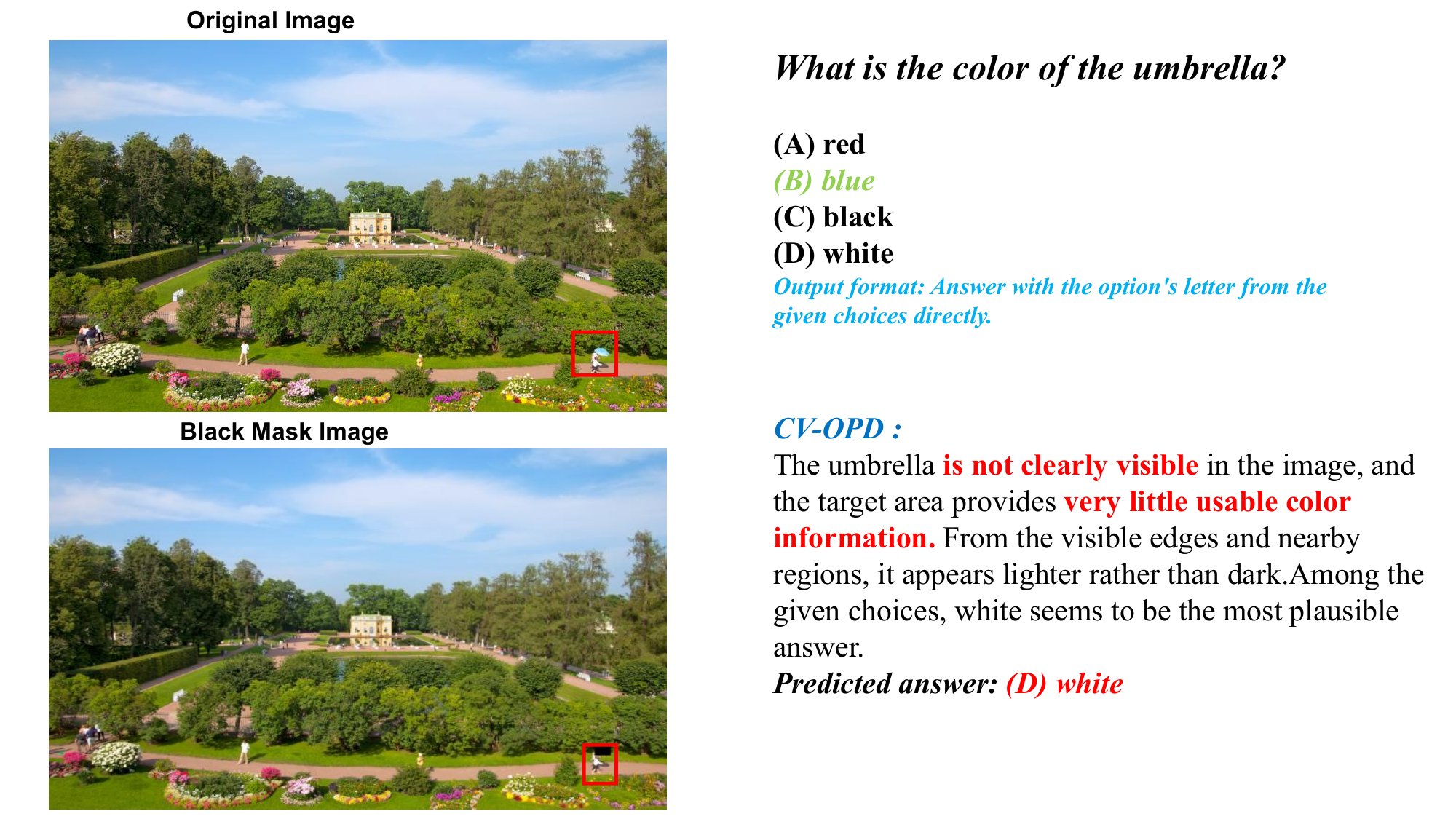}%
    }

    \endgroup

    \caption{
        CV-OPD response under the black-mask setting.
        When the umbrella is not clearly visible, CV-OPD
        explicitly acknowledges the lack of reliable color
        information and limits its description to the remaining
        visual evidence, although its forced-choice prediction
        is incorrect.
    }

    \label{fig:reject-case1}

\end{figure}


\begin{figure}[H]
    \centering

    \begingroup
    \setlength{\fboxsep}{4pt}
    \setlength{\fboxrule}{0.5pt}

    \fbox{%
        \includegraphics[
            width=0.94\linewidth,
            height=0.30\textheight,
            keepaspectratio
        ]{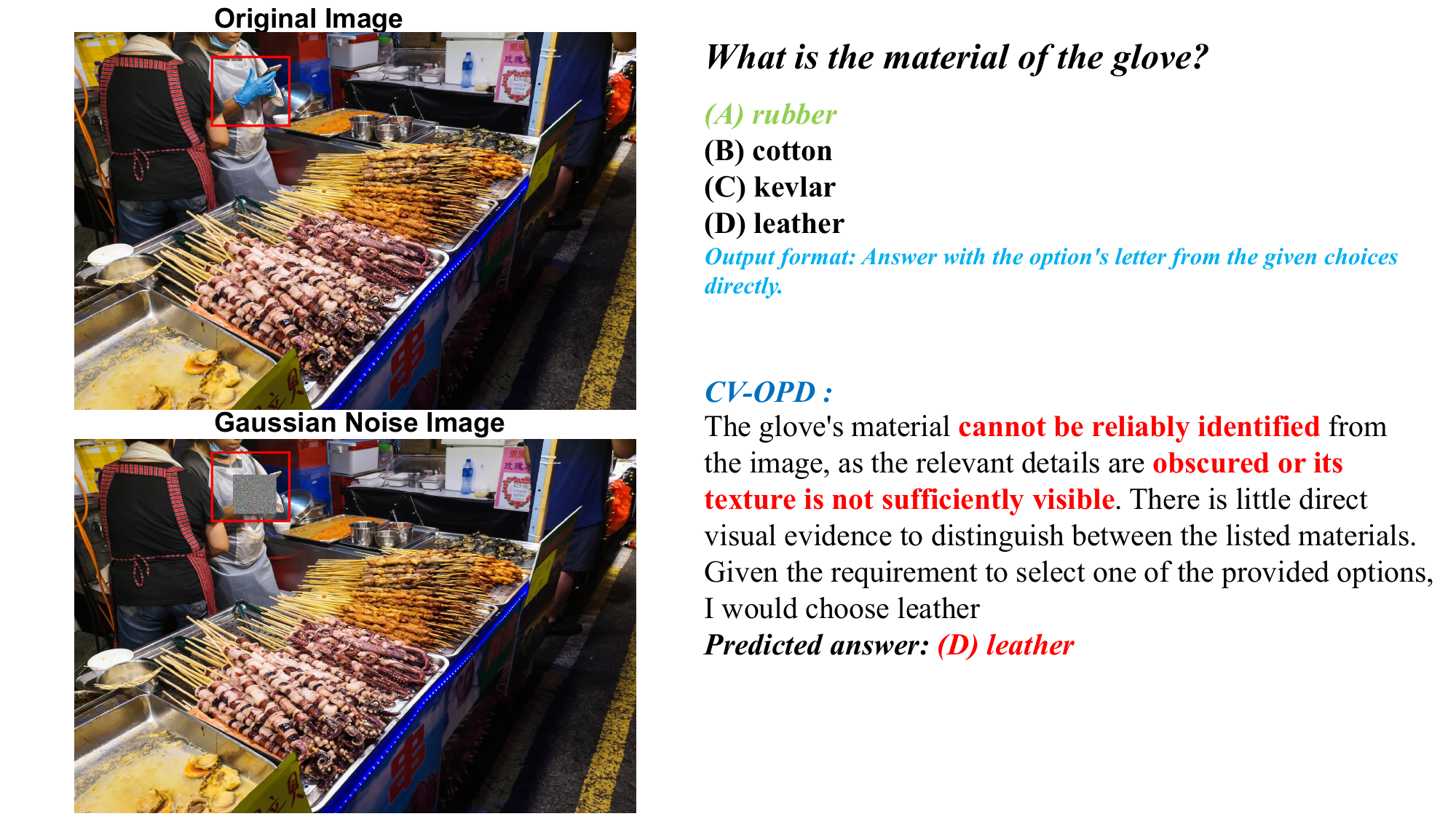}%
    }

    \endgroup

    \caption{
        CV-OPD response under Gaussian noise.
        With the glove's material obscured by visual
        corruption, CV-OPD recognizes that the available
        evidence is insufficient for reliable identification,
        despite producing an incorrect final choice.
    }

    \label{fig:reject-case2}

\end{figure}

\end{document}